\documentclass[conference]{IEEEtran}
\IEEEoverridecommandlockouts
\usepackage{cite}
\usepackage{amsmath,amssymb,amsfonts}
\usepackage{algorithmic}
\usepackage{graphicx}
\usepackage{textcomp}
\usepackage{xcolor}
\usepackage{hyperref}
\usepackage{cancel}
\usepackage{subfig}
\def\BibTeX{{\rm B\kern-.05em{\sc i\kern-.025em b}\kern-.08em
    T\kern-.1667em\lower.7ex\hbox{E}\kern-.125emX}}
\usepackage{multirow}
\usepackage{tikz}
    
\newcommand{\ie}{i.e., }
\newcommand{\eg}{e.g., }
\newcommand{\quotes}[1]{``#1''}

\begin{document}

\title{Modeling Edge Features with \\ Deep Bayesian Graph Networks \thanks{Work partially supported by SIR 2014 project LIST-IT (RBSI14STDE) and by the H2020 project TAILOR (952215).}}
\author{Anonymous Authors}
\author{\IEEEauthorblockN{Daniele Atzeni$^\ast$\thanks{$^\ast$Equal contribution.}, Davide Bacciu$^\ast$, Federico Errica$^\ast$, Alessio Micheli$^\ast$}
\IEEEauthorblockA{\textit{Department of Computer Science} \\
\textit{University of Pisa}\\
Pisa, Italy \\
daniele.atze@gmail.com, federico.errica@phd.unipi.it, \{bacciu, micheli\}@di.unipi.it}
}

\maketitle

\begin{abstract}
We propose an extension of the Contextual Graph Markov Model, a deep and probabilistic machine learning model for graphs, to model the distribution of edge features. Our approach is architectural, as we introduce an additional Bayesian network mapping edge features into discrete states to be used by the original model. In doing so, we are also able to build richer graph representations even in the absence of edge features, which is confirmed by the performance improvements on standard graph classification benchmarks. Moreover, we successfully test our proposal in a graph regression scenario where edge features are of fundamental importance, and we show that the learned edge representation provides substantial performance improvements against the original model on three link prediction tasks. By keeping the computational complexity linear in the number of edges, the proposed model is amenable to large-scale graph processing.
\end{abstract}

\begin{tikzpicture}[remember picture, overlay]
  \node [draw, rectangle, anchor=north, inner sep=5pt, text width=\textwidth, align=left] at ([yshift=-0.2cm] current page.north) {%
      \parbox{\dimexpr\linewidth-2\fboxsep}{%
      \footnotesize
        \copyright 2021 IEEE. Personal use of this material is permitted. Permission from IEEE must be obtained for all other uses, in any current or future media, including reprinting/republishing this material for advertising or promotional purposes, creating new collective works, for resale or redistribution to servers or lists, or reuse of any copyrighted component of this work in other works.\\ 
\textit{To appear in the Proceedings of the 2021 International Joint Conference on Neural Networks (IJCNN 2021)}
      }%
  };
\end{tikzpicture}


\section{Introduction}
Real-world complex systems are often modeled (and best understood) as a set of entities that interact with each other. In some cases, even in presence of a large number of empirical observations, it may be non-trivial to describe how a specific interaction system, \eg a molecule, is associated with a specific property, \eg being beneficial or harmful to humans. It is in this challenging context that the field of machine learning for graphs is experiencing an unprecedented acceleration in terms of both research \cite{micheli_neural_2009,scarselli_graph_2009} and successful applications \cite{wu_comprehensive_2019}. 

A graph is the natural mathematical abstraction we can use to model interaction systems. Entity information is stored on vertices, whereas interaction data is placed on edges. Machine learning models for graphs are capable of dealing with structures that vary in size and topology, but oftentimes the processing considers only vertex features and adjacency information \cite{bacciu_gentle_2020}. Instead, the nature of the entities' interactions, like the bond type and atomic distance in molecules, can be crucial to accurately describe the system of interest. We thus claim that it is of primary importance to investigate richer learning models which can explicitly consider the contribution of edge-related features, in addition to the vertex labels which are typically exploited by deep graph models in literature.

The purpose of this work is therefore to extend a deep probabilistic model for graphs, namely the Contextual Graph Markov Model (CGMM) \cite{bacciu_contextual_2018,bacciu_probabilistic_2020}, to the processing of arbitrary (rather than discrete) edge features. As we will see more in depth in the following sections, the benefits of the proposed methodology are multi-faceted. First and foremost, we enlarge the classes of graphs that we can handle by considering continuous (possibly multi-dimensional) edge attributes. Secondly, we show that our unsupervised model can build richer graph representations than CGMM even in the absence of edge features. Finally, we highlight how our unsupervised graph embeddings can be fed to a standard machine learning predictor to effectively tackle tasks of different nature with competitive performances. Empirically, we observe that the extended model improves the accuracy on common graph classifications benchmarks against CGMM as well as end-to-end supervised methodologies, which can be attributed to the richness of the learned graph representations. Moreover, we will compare our performances against CGMM on a graph regression and three link prediction tasks, to show the advantage that modeling the generation of edge features can give. By keeping the computational costs linear in the number of edges, the model remains scalable to very large graphs. 

The rest of the paper is organized as follows: Section \ref{sec:background} reviews the main paradigms of machine learning for graphs and the related literature; Section \ref{sec:model} describes our model and the rationales for this work in detail; Section \ref{sec:experiments} describes the experimental setup in rigorous terms; Section \ref{sec:results} reports our findings; finally, Section \ref{sec:conclusions} summarizes our work.

\section{Related Works}
\label{sec:background}
Early research in neural network-based machine learning for graphs, also known as Deep Graph Networks (DGNs) \cite{bacciu_gentle_2020}, started was pioneered by the (feed-forward) Neural Network for Graphs (NN4G) \cite{micheli_neural_2009} and the (recurrent) Graph Neural Network (GNN) \cite{scarselli_graph_2009}. In particular, the former proposed the simple and efficient neighborhood aggregation mechanism that nowadays is known as \quotes{spatial graph convolution}, other than an incremental mechanism to determine the appropriate number of layers to use. Since then, a plethora of models have been developed to tackle regression, classification and generative tasks involving vertices or entire graphs. From the neural networks paradigm, we mention: the Graph Convolutional Network \cite{kipf_semi-supervised_2017}, whose spatial convolution stems from a relaxation of spectral graph theory \cite{hammond_wavelets_2011}; GraphSAGE \cite{hamilton_inductive_2017}, which proposes an aggregation mechanism akin to that of NN4G with a sampling scheme to accelerate training and inference; and the Graph Isomorphism Network \cite{xu_how_2019}, known to be as powerful as the 1-dim WL test of graph isomorphism \cite{douglas_weisfeiler-lehman_2011} when discriminating structures.

Secondly, contributions from the field of reservoir computing, whose methods rely on a randomized parametrization of the weights, can be ascribed to the Graph Echo State Network \cite{gallicchio_graph_2010} and its recent deep version, namely the Fast and Deep Graph Neural Network \cite{gallicchio_fast_2020}. Both methods exploit the dynamics of the reservoir, built on top of the input graph, to compute embeddings for each vertex. Interestingly, these models exhibit state of the art performances, compared to neural approaches, despite the parameters of the graph convolutions being left untrained. 

Yet, only a limited portion of these models\footnote{All of which go under the name of Deep Graph Networks (DGNs) \cite{bacciu_gentle_2020}.} incorporate edge features in the graph convolution \cite{bacciu_gentle_2020}. Edge attributes are usually very informative, \eg, they represent the atomic distance between atoms or the chemical bond type. As such, being able to exploit this additional information may be critical for the downstream task.

Kernel methods \cite{kashima_marginalized_2003,borgwardt_shortest-path_2005,ralaivola_graph_2005,vishwanathan_graph_2010,shervashidze_weisfeiler-lehman_2011,yanardag_deep_2015,da_san_martino_ordered_2016,kriege_survey_2020} are yet another paradigm that has proven very effective in the domain of graphs. A kernel function for graphs takes pairs of inputs and defines a score of similarity between them. This function is used to construct a (positive semi-definite) matrix of pair-wise similarities between all the samples in the dataset, and it is usually fed into a Support Vector Machine \cite{cortes_support-vector_1995} to solve graph classification tasks. The caveat of most kernels is that, to be applicable, the kernel function must be defined in advance by human experts. In principle, this means that for each task a new and \quotes{appropriate} kernel function should be defined; in this sense, kernels are typically non-adaptive methods (although notable adaptive exceptions are seldom found in literature \cite{8259316}). Despite this, there are a number of graph classification tasks in which kernels excel at, but their applicability is often restricted to small graphs, \eg molecules in the chemical domain, due to the often impractical computational costs. Moreover, continuous edge attributes are generally not taken into account \cite{kriege_survey_2020}. In contrast, the aforementioned neural network methods are more efficient, and sometimes they perform better than kernels due to their adaptivity with respect to the task.

Until recently, no deep Bayesian methods had been developed for the efficient and flexible learning on graph-structured data. Statistical Relational Learning techniques \cite{koller_introduction_2007} usually consider tractable pair-wise dependencies between vertices in the graph, while deep neural networks take advantage of multiple layers to spread information across the graph. The Contextual Graph Markov Model \cite{bacciu_contextual_2018,bacciu_probabilistic_2020} is the first approach that brings together the probabilistic formulation adapted from tree-structured data \cite{bacciu_input_2013} and the deep incremental architecture of the NN4G to perform unsupervised learning on graphs. The model consists of a stack of Bayesian networks, trained in a layer-wise fashion to model the distribution of the vertices in the graph. 
This model produces vertex and graph representations that can be subsequently used in downstream tasks such as graph classification. An inherent limitation of CGMM, which restricts the classes of graphs it can consider, is that edge attributes must be discrete to be considered. Therefore, the purpose of our work is to address this limitation by providing an architectural solution and updated transition probabilities, so that the model can easily deal with non-discrete edge attributes.

Finally, notice that the list of methodologies presented above is far from being exhaustive due to space limitations and scope of the paper; for comprehensive surveys and introductions to the field, we refer the readers to a more appropriate literature \cite{hamilton_representation_2017,bronstein_geometric_2017,zhang_deep_2018,battaglia_relational_2018,zhang_graph_2019,wu_comprehensive_2019,bacciu_gentle_2020}.

\section{The Model}
\label{sec:model}
This section is devoted to the description of the proposed model, called E-CGMM, which extends the capabilities of CGMM to the generation of edge features. We shall give a brief overview of the necessary notation and CGMM preliminaries before delving deep into the mathematical details of the extended model.

\subsection{Mathematical Notation}
Formally, a graph is defined as a tuple $g=(\mathcal{V}_g,\mathcal{E}_g, \mathcal{X}_g, \mathcal{A}_g)$, where $\mathcal{V}_g$ is the set of vertices  representing the entities of interest, $\mathcal{E}_g$ is the set of edges connecting pairs of vertices, and $\mathcal{X}_g$ (respectively $\mathcal{A}_g$) represents the domain of vertex (respectively edge) features. For the purpose of this work, we consider edges to be directed, \ie each edge from vertex $u$ to $v$ is described as an ordered pair $(u,v)$. We denote vertex $u$ features with $\mathbf{x}_v \in \mathcal{X}_g$; similarly, edge $(u,v)$ features are associated with $\mathbf{a}_{uv} \in \mathcal{A}_g$. The neighborhood of a vertex $v$ is defined as $\mathcal{N}_v = \{u\in\mathcal{V}_g \mid (u, v)\in\mathcal{E}_g\}$. If edges hold discrete labels, we use $\mathcal{N}_v^{c} = \{u \in \mathcal{N}_v \mid a_{uv}=c\}$ to define the subset of neighbors connected to $v$ with an edge with discrete label $c$. Finally, we use $\mathcal{E}_v$ to denote the set of incoming edges in $v$, \ie $\mathcal{E}_v = \{(u, v) \in \mathcal{E}_g \mid u \in \mathcal{N}_v\}$.

\subsection{Building Blocks}
\label{subsec:building-blocks}
\begin{figure}[t]
    \centering
    \includegraphics[width=0.5\columnwidth]{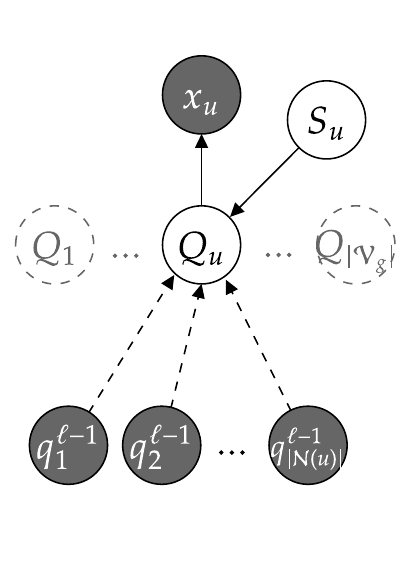}
    \caption{The graphical model associated with a generic layer $\ell$ of CGMM. Shaded vertices (respectively white) represent observable (latent) variables. The Switching Parent variable $L_u$ in \cite{bacciu_probabilistic_2020} is not used as it showed little improvement on performances at much higher computational costs. Dashed arrows denote contextual information spreading across the graph from the previous layer.}
    \label{fig:cgmm}
\end{figure}
As mentioned in Section \ref{sec:background}, CGMM is an unsupervised methodology that efficiently learns from graphs of variable topology. CGMM deep architecture is built in an incremental fashion by training one layer after another, which allows tractability of the learning equations and tackles the typical vanishing gradient problem in very deep architectures \cite{bacciu_probabilistic_2020}. Each layer $\ell$ is represented as the Bayesian network of Figure \ref{fig:cgmm}, in which a latent categorical variable $Q_u$, with $C_V$ possible states, models the generation of $\mathbf{x_u}$ conditioned on the set of $u$'s neighboring states $\mathbf{q}^{\ell-1}_{\mathcal{N}_u} = \{\mathbf{q}^{\ell-1}_{1},\dots,\mathbf{q}^{\ell-1}_{|\mathcal{N}_u|}\}$ already inferred at layer $\ell-1$. 
However, the variable number of neighboring states for each vertex makes it difficult to formalize a probability distribution with fixed parameters; moreover, the learning phase quickly becomes computationally intractable as the number of neighbors increases. 

To address these issues, CGMM resorts to the Switching Parent (SP)  \cite{saul_mixed_1999} approximation technique, which weights differently groups of neighbors sharing different \textit{discrete} edges. A new latent variable $S_u$ is therefore introduced to assign one weight to each of the resulting groups. It follows that, if $\mathcal{A}_g = \mathbb{R}^d, d \in \mathbb{N}$, we cannot apply the SP technique. This fact restricts the classes of graphs that CGMM can handle, thus hampering its performance in real-world applications where edge features encode useful information.

\begin{figure}[t]
    \centering
    \includegraphics[width=0.5\columnwidth]{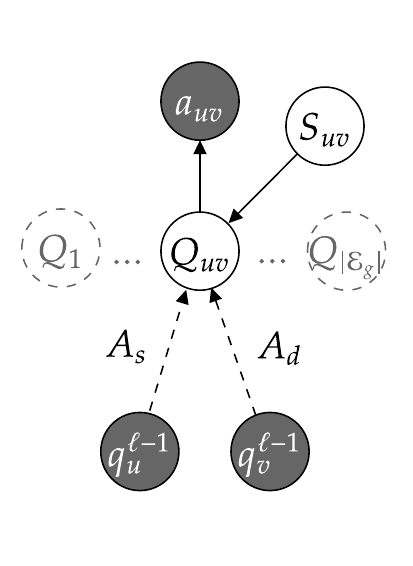}
    \caption{The graphical model of the edge-centric version of CGMM, which models the conditional distribution of edge features at layer $\ell$. By inferring latent states of the categorical variable $Q_{uv}$, we can apply the SP technique for the vertex-centric CGMM at subsequent layers of the deep architecture.}
    \label{fig:extension}
\end{figure}

We approach this problem from an architectural point of view. Rather than adding mathematical complexity to the model, we observe that the \quotes{base} CGMM model can still be used if we discretize the edge features in some way. Importantly, na\"ive discretization techniques may not work in the multidimensional setting, and these would force the user to make decisions a-priori. The idea is that we can use a learnable \textit{edge-centric} version of CGMM to discretize the edge features. This is graphically represented in Figure \ref{fig:extension}: an additional CGMM model is trained at each layer to model the generation each edge features using a latent categorical variable $Q_{uv}$ that can take $C_E$ different values. Notably, edges act as fictitious vertices whose neighbors are the source and destination vertex states inferred at the previous layer. In addition, the edge-centric CGMM uses two discrete edge labels, one for the source ($A_s$) and one for the destination $A_d$ states, to model the direction of the edge under consideration.

In summary, at the cost of training an additional network for edges that shares the same time complexity as the original CGMM (as we shall see), we are able to perform unsupervised learning on a larger class of input graphs. By training these two networks together, layer after layer, we obtain a deep architecture capable of building both vertex and edge embeddings from raw graphs, something which no other DGN can do to the best of our knowledge.

\subsection{Model Formalization}
Akin to CGMM and the neural counterparts, E-CGMM is characterized by a local and iterative processing of graph information \cite{bacciu_gentle_2020}. This means that, at layer $\ell$, each node of the graphical model considers the neighbors' contribution coming from the previous layer only. In probabilistic terms, this translates into assuming conditional independence between all vertex and edge observable variables given the corresponding latent states. Hence, we can model the generation of vertex and edge features as follows:
\begin{align}
    & P(\mathbf{x}_u) = \sum_{i=1}^{C_V} \underbrace{P_V(\mathbf{x}_u \vert Q_u=i)}_\text{vertex emission}P_V(Q_u = i \vert \mathbf{q}^{\ell-1}_{\mathcal{N}_u}, \mathbf{q}^{\ell-1}_{\mathcal{E}_u}) \label{eq:attr dist-1}
 \\
    & P(\mathbf{a}_{uv}) = \sum_{i=1}^{C_E} \underbrace{P_E(\mathbf{a}_{uv} \vert Q_{uv}=i)}_\text{edge emission}P_E(Q_{uv}=i \vert q^{\ell-1}_u, q^{\ell-1}_v), \label{eq:attr dist-2}
\end{align}
where $\mathbf{q}^{\ell-1}_{\mathcal{N}_u}$ and $\mathbf{q}^{\ell-1}_{\mathcal{E}_u}$ denote the set of states inferred by the vertex and edge components of the previous layer, respectively. Also, note how we introduced the latent variables thanks to the usual marginalization technique. Note that, when $\ell=0$, the equations simplify and the layer implements a standard mixture model that does not consider the graph structure.

The conditional distributions appearing in the rightmost part of Eq. \ref{eq:attr dist-1} and \ref{eq:attr dist-2} are the cause of the aforementioned computational intractability. We start addressing the latter, by introducing an SP variable $S_{uv}$ that weights the contributions of the source and destination vertices. Hence, we approximate the conditional distribution of the latent edge variables as
\begin{align}
\label{eq:edge post}
    & P_E(Q_{uv}=i\vert q^{\ell-1}_u, q^{\ell-1}_v) = \nonumber \\
    & = \sum_a^{A_s, A_d} \underbrace{P_E(S_{uv} = a)}_{SP_E}\underbrace{P^{a}_E(Q_{uv}=i\vert q^{\ell-1}_{a})}_\text{edge transition}
\end{align}
where we remind that $A_s$ and $A_d$ are the discrete labels assigned to source and destination vertices, respectively, and $q^{\ell-1}_{a}$ is $q^{\ell-1}_u$ if $a=A_s$, and $q^{\ell-1}_v$ otherwise.
As for Eq. \ref{eq:attr dist-1}, we apply the same approximation in CGMM, but with the notable difference that discrete edge labels are given by the \textit{edge states} inferred at the previous layer:
\begin{align}
\label{eq:node post}
    & P_V(Q_u=i \vert \mathbf{q}^{\ell-1}_{\mathcal{N}_u}, \mathbf{q}^{\ell-1}_{\mathcal{E}_u}) = \nonumber \\
    & = \sum_{a=1}^{C_E}\underbrace{P_V(S_u=a)}_{SP_V}\underbrace{P^{a}_V(Q_u=i\vert \mathbf{q}^{\ell-1}_{\mathcal{N}^a_u})}_\text{vertex transition},
\end{align}
where $\mathbf{q}^{\ell-1}_{\mathcal{N}^a_u}$ is the subset of vertices in $\mathbf{q}^{\ell-1}_{\mathcal{N}_u}$ whose edges have (discrete) label $a$. This last equation, while apparently similar to the one of CGMM (see Appendix A of \cite{bacciu_probabilistic_2020}), shows the interplay between the vertex-centric and edge-centric components of E-CGMM, which makes it possible to incorporate arbitrary edge information in the graph convolution. This characteristic is at the core of E-CGMM: apart from extending the processing to a larger class of graphs, the proposed architectural extension allows for the computation of meaningful edge states even when an edge feature is \textit{missing} (with a constant value on $\mathbf{a}_{uv}$).

The last brick in the formalization of the model is the definition of the transition distribution $P^{a}_V(Q_u=i\vert \mathbf{q}^{\ell-1}_{\mathcal{N}^a_u})$ of Eq. \ref{eq:node post}, which is still intractable due to the variable number of contributions in the set $\mathbf{q}^{\ell-1}_{\mathcal{N}^a_u}$. Using the additional edge information with respect to CGMM, we can write
\begin{align}
    & P^{a}_V(Q_u=i \vert \mathbf{q}^{\ell-1}_{\mathcal{N}^a_u}) = \frac{1}{\sum_{v\in\mathcal{N}_u}q_{uv}^{\ell-1}(a)} \times \nonumber \\
    & \times \sum_{j=1}^{C_V}P_V(Q=i \vert q=j)\sum_{v\in\mathcal{N}_u}q_v^{\ell-1}(j)q_{uv}^{\ell-1}(a)
\label{eq:node-aggr}
\end{align}
where $q_u(j)$ and $q_{uv}(j)$ are the $j$-th components of the posterior distributions (represented as a vector) inferred at the previous layer. The transition distribution of Eq. \ref{eq:node-aggr} is just a generalization of the one defined in \cite{bacciu_probabilistic_2020}, but it exploits the posterior probability vector of each edge to weight the contribution of individual neighbors. Finally, as in CGMM, we assume full stationarity on vertices and on edges, meaning that we share the parameters of the emission, transition, and SP distributions across all vertices or edges depending on the component of E-CGMM. This allows us to deal with graphs of different sizes and reduces overparametrization.

\subsection{Dynamic Vertex Aggregation}
\label{subsec:dynamic}
\begin{figure}[t]
    \centering
    \includegraphics[width=\columnwidth]{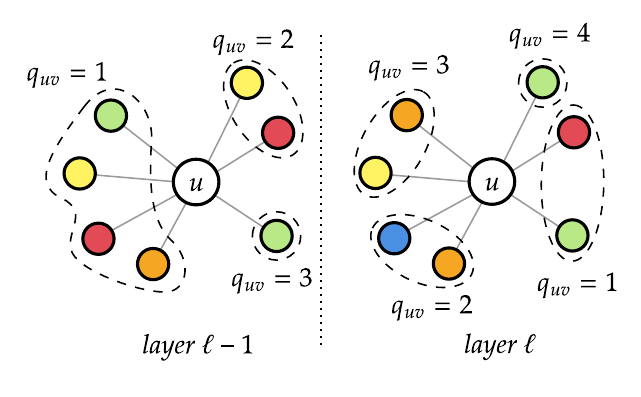}
    \caption{We show an example of dynamic neighborhood aggregation with $C_E=4$. Colors denote different vertex states in $\mathbf{q}^{\ell-1}_{\mathcal{N}_u}$. At layer $\ell-1$, the neighbors of vertex $u$ are split into 3 groups according to the edge states computed at layer $\ell-2$. Because edge states may vary in E-CGMM, at layer $\ell$ a different grouping of the same neighbors is possible, according to the edge states inferred at layer $\ell-1$.} 
    \label{fig:dynamic-aggr}
\end{figure}

There is one subtle difference between CGMM and E-CGMM that potentially contributes to the richness of the vertex embeddings produced by the latter. Even when edge features are missing, the edge latent variables can still vary across the graph, as they depend on the source and destination states of Eq. \ref{eq:edge post}. This means that, in principle, at each layer the posterior distribution of an edge may have different values regardless of the presence or absence of an edge feature. In turn, this allows for different grouping of the same vertex neighbors at different layers. In contrast, CGMM would always group neighbors in the same way. This \quotes{dynamic neighborhood aggregation} is sketched in Figure \ref{fig:dynamic-aggr}. We believe this is the main reason why E-CGMM shows (in our empirical analysis) significant performance improvements with respect to CGMM on classical benchmarks.

\subsection{Training}
To train each layer of E-CGMM, we frame learning as a maximum likelihood estimation problem. Because the vertex and edge Bayesian networks at layer $\ell$ are always independent from each other, we can maximize the sum of the individual likelihoods. By considering a set of i.i.d. graphs $\mathcal{G}$ at a given layer $\ell$, we formulate the vertex and edge likelihoods as:
\begin{align}
    & \mathcal{L}_V(\theta \vert \mathcal{G}) = \prod_{\substack{g\in\mathcal{G} \\ u\in\mathcal{V}_g}}\sum_{i=1}^{C_V} P(\mathbf{x}_u \vert Q_u=i)P(Q_u \vert \mathbf{q}^{\ell-1}_{\mathcal{N}_u}, \mathbf{q}^{\ell-1}_{\mathcal{E}_u}) \\
    & \mathcal{L}_E(\theta \vert \mathcal{G}) = \prod_{\substack{g\in\mathcal{G} \\ uv\in\mathcal{E}_g}}\sum_{i=1}^{C_E} P(\mathbf{a}_{uv} \vert Q_{uv}=i)P(Q_{uv} \vert q^{\ell-1}_{u}, q^{\ell-1}_{v}).
\end{align}
Each likelihood is then maximized via the Expectation Maximization (EM) \cite{dempster_maximum_1977} algorithm, which provides convenient convergence guarantees. In what follows, we describe the most general formulation associated with the vertex component of Figure \ref{fig:cgmm}; indeed, the generative model of Figure \ref{fig:extension} is a \quotes{restricted} version of the former. 

To compute the E-step of the EM algorithm, we introduce the indicator variable $z_{uiaj}$, which assumes value $1$ if vertex $u$ is in state $i$ while its neighbors with edge state $a$ are in state $j$ at the previous layer, and $0$ otherwise. Then, the required expected value of the indicator variable can be computed as
\begin{align}
    & E[z_{uiaj}\vert\mathcal{G}, \mathbf{q}] = P(Q_u=i, S_u=a, q=j\vert\mathcal{G}, \mathbf{q}) \nonumber\\
    & = \frac{1}{Z}P(x_u\vert Q_u=i)P(S_u=a)P^a(Q_u=i\vert q=j),
\end{align}
where $Z$ is the usual normalization term obtained by straightforward marginalization. By marginalization over the indices of the indicator variable $z_{uiaj}$, we obtain the posterior probabilities used to update the model parameters at the M-step:
\begin{align}
    & P(S=a) = \frac{1}{Z}\sum_{\substack{g\in\mathcal{G} \\  u\in\mathcal{V}_g \\ i\in\{1,\dots,C_V\}}} E[z_{uia} \vert \mathcal{G}, \mathbf{q}] \\
    & P^a(Q=i \vert q=j) = \frac{1}{Z}\sum_{\substack{g\in\mathcal{G} \\ u\in\mathcal{V}_g}} E[z_{uiaj} \vert \mathcal{G}, \mathbf{q}].
\end{align}
The update rule for the emission distribution depends on whether we are dealing with discrete or continuous features. We exemplify the M-step update for the categorical distribution, but the reader can refer to \cite{bacciu_probabilistic_2020} for the Gaussian case
\begin{equation}
    P(y=k \vert Q=i) = \frac{1}{Z} \sum_{\substack{g\in\mathcal{G} \\ u\in\mathcal{V}_g}}\delta (x_u, k) E[z_{ui} \vert \mathcal{G}, \mathbf{q}]
\end{equation}
where $\delta (\cdot, \cdot)$ is the Kronecker delta. 

\subsection{Inference}
\label{subsec:inference}
The inference phase computes the most likely latent state for each vertex or edge. To do that, for a generic layer $l$ and vertex $u$, we compute posterior probabilities by means of the Bayes' theorem
\begin{align}
    & \max_{i}P(Q_u=i \vert g, \mathbf{q}^{\ell-1}_{\mathcal{N}_u}, \mathbf{q}^{\ell-1}_{\mathcal{E}_u}) = \nonumber \\
    & = \frac{P(\mathbf{x}_u \vert Q_u=i)P(Q_u \vert \mathbf{q}^{\ell-1}_{\mathcal{N}_u}, \mathbf{q}^{\ell-1}_{\mathcal{E}_u})}{\cancel{P(\mathbf{x_u}\vert\mathbf{q}^{\ell-1}_{\mathcal{N}_u}, \mathbf{q}^{\ell-1}_{\mathcal{E}_u})}},
\end{align}
where the constant denominator does not count in the maximization. Inference for each edge is obtained similarly
\begin{align}
    & \max_{i}P(Q_{uv}=i \vert g, q^{\ell-1}_{u}, q^{\ell-1}_{v}) = \nonumber \\
    & = \frac{P(\mathbf{a}_{uv} \vert Q_{uv}=i)P(Q_{uv} \vert q^{\ell-1}_{u}, q^{\ell-1}_{v})}{\cancel{P(\mathbf{a
    }_{uv}\vert q^{\ell-1}_{u}, q^{\ell-1}_{v})}}.
\end{align}

\subsection{Embeddings Construction}
As we just discussed, at the end of the inference phase we obtain a vector of posterior probabilities from which we can extract the most likely hidden state for $Q_u$ and $Q_{uv}$. However, when computing a vertex or edge frozen states, we can decide to use a one-hot encoding of the most likely state or to simply use the posterior vector of probabilities. This decision between the discretized and continuous representations of states is left as a hyper-parameter of the model. In both cases, we refer to the vertex (respectively edge) representation of size $C_V$ ($C_E$) with the term \textit{unigram}. Following \cite{bacciu_probabilistic_2020}, we can also augment vertices representations with vertex \textit{bigram} information, \ie a $C_V^2$-sized vector containing statistics about the neighborhing states. Formally, the bigram of a vertex $u$ is defined as
\begin{align}
    & \phi_{i_j}(u) = \sum_{v\in \mathcal{N}_u} q_u(i)q_v(j) \quad i,j=1,\dots, C_V .
\end{align}
At model selection time, we can then choose to concatenate a vertex unigram and bigram (obtaining a \textit{unibigram}) or not. To obtain graph representations at each layer $\ell$, we independently aggregate all vertex and edge representations in the graph via permutation-invariant operators, such as the mean or the sum, and concatenate the two resulting vectors. The final unsupervised graph embedding is then the concatenation of the embeddings across the layers of the deep architecture. This graph embeddings will be used to train a predictor on the tasks described in the next section.

\subsection{Computational Considerations}
We now show that E-CGMM shares comparable asymptotic efficiency in time and space to CGMM. At a given layer $\ell$, the complexity of E-CGMM vertex component, \ie CGMM, can be bounded in time and space by $\mathcal{O}(\vert\mathcal{V}_g\vert \times C_E \times C_V^2)$, which is basically the cost of learning the transition and emission distributions. To compute the posteriors and bigram of each vertex, the time complexity is  $\mathcal{O}(\vert\mathcal{E}_g\vert)$, which is the cost of access to the structure. Hence, asymptotically, the overall time complexity is $\mathcal{O}(\vert\mathcal{V}_g\vert + \vert\mathcal{E}_g\vert)$. Instead, the edge component of E-CGMM is bounded by $\mathcal{O}(\vert\mathcal{E}_g\vert \times 2 \times C_E \times C_V)$. Clearly, the time complexity of our extension will be always strictly greater than CGMM; however, asymptotically speaking, it is still controlled by $\mathcal{O}(\vert\mathcal{V}_g\vert + \vert\mathcal{E}_g\vert)$ whenever $C_V \ll \vert\mathcal{E}_g\vert, C_E \ll \vert\mathcal{E}_g\vert$, thus making the model amenable to large-scale graph processing.

\section{Experiments}
\label{sec:experiments}
In this section, we describe in detail the experimental setting used in order to reproduce our results. 
We carry out an empirical evaluation of E-CGMM against three different types of tasks. First, we compare our model on standard graph classification benchmarks. Secondly, we tackle a chemical graph regression task where continuous edge features are available. 
Finally, we test the effectiveness of E-CGMM against CGMM on three link prediction tasks\footnote{Code for reproducibility: \url{https://github.com/diningphil/E-CGMM}}.

\subsection{Graph Classification}
We consider two popular graph classification benchmarks, namely NCI1 \cite{nci1} and COLLAB \cite{collab}. The first is a chemical dataset of more than 4000 compounds, where 
the task is to classify each compound according to its ability to suppress or not the growth of cancer cells. The second is a scientific collaboration dataset 
that counts 5000 social networks. Here, the task is to classify the specific field each network belongs to. Both datasets do not provide edge attributes, so the scope of our analysis is to check the richness of graph embeddings given by the dynamic aggregation mechanism of Section \ref{subsec:dynamic} and the use of posterior edge probabilities conditioned on the pair of source and destination states. Following \cite{bacciu_probabilistic_2020}, we use the vertex degrees as continuous vertex features on COLLAB, whereas NCI1 discrete vertex features represent atoms.


We perform 10-fold cross validation for model assessment as in \cite{bacciu_probabilistic_2020}, with an hold-out technique for model selection inside each fold. 
In terms of hyper-parameters, we set $C_V$ to 20 and the EM iterations to $20$ (respectively $10$) for NCI1 (COLLAB). $C_E$ is chosen in $\{5, 10\}$, the number of layers in $\{10, 20\}$ for NCI1 and in $\{10, 15, 20\}$ for COLLAB. We experiment with both unigram and unibigram version of graph embeddings with mean aggregation, and hidden state representations can be either discrete or continuous. Graph embeddings serve as input for an MLP with one hidden layer and ReLU as activation function, trained with Adam \cite{kingma_adam_2015} optimizer on Cross Entropy. For NCI1, we set the maximum number of epochs to $2000$, the hidden layer dimension to $128$, and the learning rate to $5\cdot 10^{-4}$. For COLLAB, we set the epochs to $5000$ and we chose the hidden layer dimension in $\{32, 128\}$, and the learning rate in  $\{10^{-3}, 10^{-4}\}$. We also use the early stopping technique \cite{prechelt_early_1998} with patience set to $100$ for NCI1 and $500$ for COLLAB, and try the $L2$ regularization with weight decay equal to $10^{-4}$ for NCI1 and $\{0., 5\cdot10^{-4}\}$ for COLLAB. We compare our model against 
end-to-end supervised deep learning models for graphs. Please refer to \cite{bacciu_probabilistic_2020} for a description of these methods.

\subsection{Graph Regression}
To understand the importance of handling continuous edge information with no a-priori discretization, we consider the QM7b graph regression task \cite{qm7b, qm7b2}, a chemical dataset composed of more than 7k organic molecules. The task is to predict $14$ continuous properties of each molecule, where each molecule is associated with a Coulomb matrix that is used to extract the vertex and edge features. 
The 6 different diagonal elements of the matrix are used as discrete labels for the vertices, while we consider edges associated with matrix entries greather than $0.52$ to induce sparsity on the graph. 
To show that edge discretization techniques may fail, we consider a version of the dataset where we discretize the continuous features using $10$ bins of equal widths, so that CGMM can be trained on it. 
We use the same model selection/assessment setting of graph classification. We try configurations with depth in $\{10, 20\}$, $C_E$ in $\{5, 10\}$, $C_N$ in $\{10, 20\}$ for CGMM and $20$ for E-CGMM, continuous or discrete frozen states, unigrams or unibigrams, and sum or mean aggregation. The MLP configurations are the same of NCI1, but with Mean Absolute Error (MAE) as loss/score function.

\subsection{Link Prediction}
We perform link prediction on Cora, Citeseer \cite{citeseer_cora}, and Pubmed \cite{pubmed}, three different citation networks in which vertices represent documents, and edges represent citations. The goal of these experiments is to show how E-CGMM can deal with link prediction much better than CGMM.

In this experiment, 
a portion of positive edges in each graph is used as validation or test set, together with a randomly selected subset of negative edges. The remaining edges are used to train E-CGMM. Since CGMM doesn't return edge embeddings, we use an MLP to predict whether a link $(u,v)$ exists by using the mean embedding of vertices $u$ and $v$. In particular, we use the mean rather concatenation because the latter would introduce asymmetries (and therefore learning difficulties) in modeling undirected edges. 
Once E-CGMM is trained, we infer a link's label as the mean predicted label across all layers. 
For every dataset and for both CGMM models, we choose in model selection: $C_V \in \{10, 20\}$, model depth in $\{2, 4, 6, \dots , 20\}$, learning rate in $\{10^{-3}, 10^{-4}, 10^{-5}\}$, hidden dimension in $\{128, 256\}$ and weight decay in  $\{10^{-3}, 10^{-5}\}$. We also try both the continuous and discrete versions of the embeddings and, for E-CGMM, $C_E \in \{5, 10\}$.

\section{Results}
\label{sec:results}
Results for graph classification are shown in Table \ref{tab:graph_class}.
We can see that E-CGMM performs better than the related methods from literature tested on both datasets. This result shows that our unsupervised model can produce rich representations regardless of the downstream task they are applied to, even when compared with supervised models trained in an end-to-end fashion. 
\begin{table}[t]
    \centering
    \begin{tabular}{lc|c c}
    \hline
    & Model            & NCI1                     & COLLAB \\
    \hline
    & PSCN \cite{patchysan} & $76.34\pm1.7$ & $72.60\pm2.15$ \\
    & DCNN \cite{dcnn} & $56.61\pm1.0$ & $52.11\pm0.7$ \\
    & ECC \cite{ecc}  & $76.82$ & - \\
    & DGK \cite{deep-graphlet-kernel} & $62.48\pm0.3$ & $73.09\pm0.3$ \\
    & DGCNN \cite{dgcnn} & $74.44\pm0.5$ & $73.76\pm0.5$ \\
    & PGC-DGCNN \cite{filtersizegraphconvnet} & $76.13\pm0.7$ & $75.00\pm0.58$ \\ 
    \hline
    & CGMM \cite{bacciu_probabilistic_2020}             & $77.80\pm1.9$            & $75.50\pm2.74$ \\ 
    & E-CGMM           & $\mathbf{79.22} \pm 1.6$ & $\mathbf{77.80} \pm 2.1$ \\
    \end{tabular}
    \vskip 0.1in
    \caption{E-CGMM’s results for graph classification. Best results are reported in bold.} 
    \label{tab:graph_class}
\end{table}
We can also note significant improvements with respect to CGMM. We believe this is attributable to the new capabilities introduced with the edge component of each layer, \ie the ability to extract edge information by considering pairs of frozen states, and the dynamic aggregation of neighbors described in Section \ref{subsec:dynamic}. In fact, the first allows 
to incorporate 
knowledge about the pairs of vertex hidden states connected in each graph, whereas the second makes the model more flexible than CGMM by weighting differently the neighbors' contribution at each layer.

\begin{figure}[t]
    \centering
    \includegraphics[width=0.8\columnwidth]{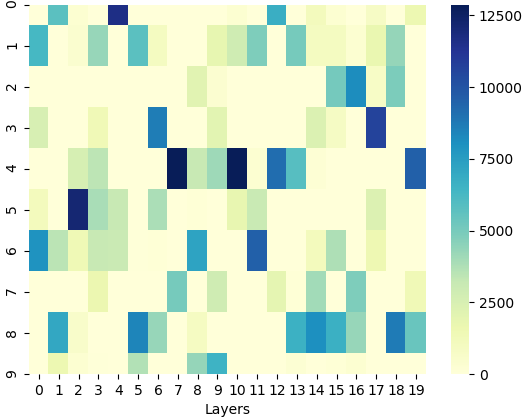}
    \caption{Heatmap representing the number of edges’ hidden state in each layer, with 10 possible hidden states on COLLAB. The color of the cell $(i, j)$ represents the number of edges whose hidden state is $i$ at layer $j$.}
    \label{fig:hidden_states}
\end{figure}

\begin{figure}[t]
    \centering
    \includegraphics[width=0.8\columnwidth]{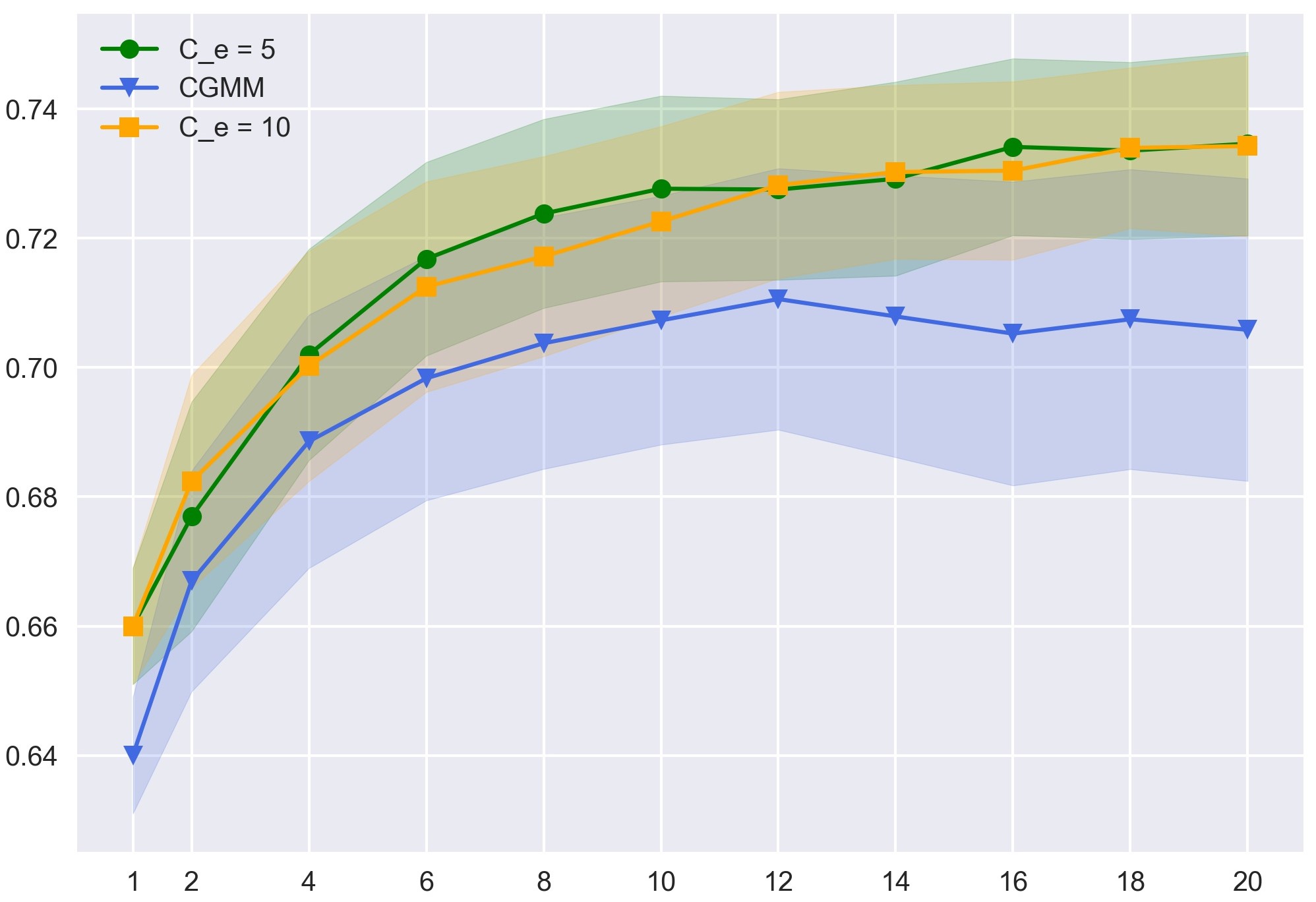}
    \caption{Average validation accuracy on NCI1 among all model configurations. Each point is obtained by grouping accuracy by the depth of the architecture and the number of possible edge hidden states.}
    \label{fig:NCI1_comp}
\end{figure}

To further investigate the importance of edge hidden states, we report qualitative results in Figure \ref{fig:hidden_states} and \ref{fig:NCI1_comp}. The first is a heatmap of edge states obtained after a training run on the COLLAB dataset. The color of the cell $(i, j)$ in Figure \ref{fig:hidden_states} represents the number of edges in state $i$ at layer $j$. We observe that, even without edge features, E-CGMM has the ability to produce different edge states. This shows how our model can produce richer graph embeddings than CGMM by means of the dynamic neighborhood aggregation mechanism.

The second Figure shows a comparison between E-CGMM and CGMM during different configuration runs on NCI1. Each data point in the figure is obtained as the average of the validation accuracy over all configurations tried during model selection, grouped by the number of possible edge hidden states and the depth of the architecture. The Figure clearly shows an average improvement independent from the number of layers. Also, a larger $C_E$ 
does not seem particularly relevant on NCI1: this is explained by the fact that, in our experiments on NCI1, edges could assume around $3/4$ different states. 
In this respect, an interesting future line of research could the the automatization of the choice of $C_E$ via Bayesian nonparametric methods. 
Similarly to \cite{bacciu_probabilistic_2020}, the accuracy is an increasing function of the depth, which stresses the importance of deep architectures in spreading information across the graph. This observation represents an additional motivation for the incremental construction of the architecture. In fact, (E-)CGMM allows to stack a large number of layers without facing the typical deep learning problem of the vanishing of the gradient \cite{deeper-insights-gcn}. It could also lead to the removal of the decision a-priori on the number of layers, since one can decide to automatically stop the construction of the architecture with some criterion, \eg based on the validation score or information-theoretic properties of the inferred states. We leave these appealing challenges for future research.

Table \ref{tab:graph_regr} reports graph regression results. We can see that E-CGMM performs better than both versions of CGMM, which was to be expected given the added flexibility of the model in dealing with continuous edge features. Such an improvement also proves the inadequacy of non-adaptive edge discretization techniques, which not only determine the necessity to take decisions a-priori but also cause a loss of relevant information.

\begin{table}[t]
    \centering
    \begin{tabular}{c | c  c} 
     \hline
     Model & MAE & Relative Improvement\\ 
     \hline
     CGMM - no edge attributes & $1.52 \pm 0.05$ & $19\%$ \\
     CGMM - discretized edges   & $1.49 \pm 0.07$ & $17\%$\\ \hline
     E-CGMM & $\mathbf{1.23} \pm 0.06$ & - \\
     \hline
    \end{tabular}
    \vskip 0.1in
    \caption{Graph regression results and relative improvement of E-CGMM compared to CGMM. Best results are in bold. CGMM results are reported for both a version of the dataset with no edge attributes as well as for discretized edge labels.}
    \label{tab:graph_regr}
\end{table}
As for link prediction, Table \ref{tab:link_pred} reports significant improvements with respect to CGMM in every dataset tested, with an increase in average accuracy of 3-4 points. By modeling the generation of positive and negative edges, E-CGMM captures the conditional distribution of the edges given the frozen vertex states, thus building more informative edge posteriors. On the other hand, the aggregation of node representations by CGMM is another a-priori choice that we aim to avoid with graph representation learning methods. 
\begin{table}[h]
    \centering
    \begin{tabular}{c | c  c  c} 
     \hline
     Model & Cora & Citeseer & Pubmed \\ 
     \hline
     CGMM & $82.62 \pm 1.8$ & $74.47 \pm 2.2$ & $77.09 \pm 1.9$ \\
     E-CGMM & $\mathbf{86.76} \pm 2.3$ & $\mathbf{77.69} \pm 1.7$ & $\mathbf{81.58} \pm 1.6$ \\
     \hline
    \end{tabular}
    \vskip 0.1in
    \caption{Comparison between E-CGMM and CGMM on link prediction tasks. Best results are reported in bold.}
    \label{tab:link_pred}
\end{table}

\section{Conclusions}
\label{sec:conclusions}
We have introduced an extended version of the Contextual Graph Markov Model that allows the processing of a broader class of graphs with labelled edges. Our proposal builds upon the introduction of an additional Bayesian network that is responsible for the generative modeling of edge features. By means of a dynamic neighborhood aggregation scheme, we obtain state of the art results on common graph classification benchmarks that do not rely on edge features. Moreover, we demonstrate the value of modeling the additional edge information on graph regression and link prediction tasks, obtaining substantial performance improvements with respect to the original model. The asymptotic complexity of the model remains linear in the number of edges, and as such the model is still applicable to large-scale graph learning. 

\bibliographystyle{ieeetr}
\bibliography{main.bib}

\end{document}